\pdfoutput=1

\documentclass[11pt]{article}

\usepackage[final]{acl}

\usepackage{times}
\usepackage{latexsym}

\usepackage[T1]{fontenc}

\usepackage[utf8]{inputenc}

\usepackage{microtype}

\usepackage{inconsolata}

\usepackage{graphicx}

\usepackage{amsthm}
\newtheorem{definition}{Definition}

\newcommand{\comment}[1]{}

\title{Controllable Synthetic Clinical Note Generation with Privacy Guarantees}

\author{
  Tal Baumel$^{1}$, Andre Manoel$^{2}$, Daniel Jones$^{1}$,  Shize Su$^{1}$, \\
  \textbf{Huseyin Inan$^{1}$, Aaron (Ari) Bornstein$^{1}$ and Robert Sim$^{1}$} \\
  $^1$Microsoft Corporation \\
  $^2$Gretel.ai (Work done while at Microsoft)\\
  \texttt{\{tabaumel, shizs, jonesdaniel, huseyininan,}\\
  \texttt{aribornstein, rsim\}@microsoft.com} \texttt{andremanoel@gmail.com} \\
}

\begin{document}
\maketitle
\begin{abstract}
In the field of machine learning, domain-specific annotated data is an invaluable resource for training effective models. However, in the medical domain, this data often includes Personal Health Information (PHI), raising significant privacy concerns. The stringent regulations surrounding PHI limit the availability and sharing of medical datasets, which poses a substantial challenge for researchers and practitioners aiming to develop advanced machine learning models. In this paper, we introduce a novel method to "clone" datasets containing PHI. Our approach ensures that the cloned datasets retain the essential characteristics and utility of the original data without compromising patient privacy. By leveraging differential-privacy techniques and a novel fine-tuning task, our method produces datasets that are free from identifiable information while preserving the statistical properties necessary for model training. We conduct utility testing to evaluate the performance of machine learning models trained on the cloned datasets. The results demonstrate that our cloned datasets not only uphold privacy standards but also enhance model performance compared to those trained on traditional anonymized datasets. This work offers a viable solution for the ethical and effective utilization of sensitive medical data in machine learning, facilitating progress in medical research and the development of robust predictive models.

\end{abstract}

\section{Introduction}

Training Healthcare Language Models for downstream tasks requires quality clinical data that represents the real-world distribution of clinical texts. Acquiring such clinical data at scale is expensive and comes with many restrictions to protect patient privacy including siloing (data cannot leave the boundaries of health care providers), de-identification (Personal Identifiable Information (PII) and PHI needs to be removed), and data retention restrictions (patients and health care providers reserve the right to revoke access to training data at any time).
Data retention restrictions prevent the reproducibility of machine learning methods, as well as limit the value of expensive annotation of this data. We propose a novel method for generating controlled synthetic clinical data leveraging semantically informed instruction tuning datasets that preserve the  style of siloed health care data and can be used for training Healthcare LLMs while preserving clinical accuracy, data privacy, and downstream modeling performance without violating data retention restrictions.
To resolve the healthcare data retention limitations generative models must generate outputs that meet the following conditions. 
Output must be:
\begin{enumerate}
    \item Stylistically akin to the original data.
    \item Clinically accurate.
    \item Privacy Preserving of the patients mentioned and not expose PII/PHI from the original data.
    \item Diverse enough to be useful for modeling downstream tasks.
\end{enumerate}

In this paper we introduce a process that decouples the limitations of data retention while preserving if not improving, patient privacy.
By creating synthetic clinical notes in the style of a given dataset users can freely use the generated documents without any limitations. For example, if we are limited to a year of use there is a strong incentive to not invest in annotating the data since the annotations will have no utility once the terms of use expire. By generating synthetic data, we can annotate it without time restrictions and use the data freely for model training.
The process can be used in the future to create an expert language model trained on data from a specific domain, that was never available before without manually annotating massive amounts of data with instructions.

\section{Related Work}
\subsection{Differential Privacy}
Our approach employs differential privacy which enables model training that is relatively invariant to the presence or absence of a single patient. The following is a formal definition of differential privacy:
\begin{definition}[Differential Privacy (DP) \cite{DworkMNS06,DworkKMMN06}]
  A randomized algorithm $\mathcal{A}$ is  ($\epsilon$,$\delta$)-differentially private if for any two neighboring datasets $D$ and $D'$, which differ in exactly the data pertaining to a single user, and for all sets $\mathcal{S}$ of possible outputs: 
$
\textstyle{\Pr[\mathcal{A}(D) \in \mathcal{S}] \leq e^{\epsilon}\Pr[\mathcal{A}(D') \in \mathcal{S}] +\delta}
$.
\end{definition}

\subsection{Differentially Private Stochastic Gradient Descent (DP-SGD)}
Differentially Private Stochastic Gradient Descent (DP-SGD)~\citep{SongCS13, BassilyST14, AbadiCGMMTZ16} is a variant of the traditional SGD algorithm, tailored to ensure differential privacy during the training process. The key idea is to add controlled noise to the gradients, thereby masking the contribution of any single data point.

The DP-SGD algorithm works as follows:
\begin{enumerate}
  \item \textbf{Gradient Computation}: Compute the gradient of the loss function with respect to each data point in a randomly selected mini-batch.
  \item \textbf{Gradient Clipping}: Clip the gradients to ensure they have bounded norm. This prevents any single gradient from having too much influence.
  \item \textbf{Noise Addition}: Add noise, typically drawn from a Gaussian distribution, to the clipped gradients. The scale of the noise is calibrated to the desired level of privacy, controlled by the parameters $\epsilon$ and $\delta$.
  \item \textbf{Parameter Update}: Update the model parameters using the noisy gradients.
\end{enumerate}

\subsection{Training Language Models with Differential Privacy}
Differential privacy can be leveraged to train language models in a manner that protects the privacy of the underlying training data. When training a language model, the goal is to learn the statistical properties of the language without memorizing specific details about any individual data point.

Using DP-SGD, we can ensure that the model updates do not inadvertently leak information about any single training example. This is particularly important when training on sensitive datasets, such as those containing Personal Health Information or proprietary text.

\comment{
\subsection{Creating Privacy-Preserving Datasets}
By leveraging differential-privacy techniques and novel fine-tuning tasks, we can produce datasets that are free from identifiable information while preserving the statistical properties necessary for model training. The process includes:
\begin{itemize}
  \item \textbf{Data Sanitization}: Apply techniques such as data anonymization and noise addition to sanitize the dataset before training. 
  \item \textbf{Fine-Tuning Tasks}: Use carefully designed fine-tuning tasks that focus on learning general patterns in the data rather than memorizing specific examples.
  \item \textbf{Privacy Audits}: Conduct thorough audits of the dataset and the trained models to ensure compliance with privacy standards and to identify any potential leaks of sensitive information.
\end{itemize}

Through these methods, we aim to achieve a balance between data utility and privacy, allowing for the effective training of language models \cite{yu2021differentially} without compromising the privacy of the individuals represented in the training data.
}

\subsection{Clinical Structuring of Unstructured Text}
In this work, we extract structure using Text Analytics for Health (TA4H)\cite{ta4hblog}. TA4H is a tool for extracting clinical named entities (e.g., medication, symptoms), linking the entities to anthologies such as UMLS \cite{bodenreider2004unified}, relationships between entities (e.g., medication dosage, body part direction), and assertions (e.g., negation, temporality).

\subsection{Automatically Generated Instruction Datasets}
\begin{figure}[t]
    \centering
    \includegraphics[width=0.48\textwidth]{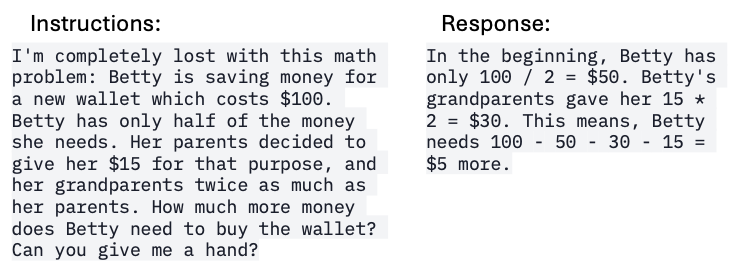}
    \caption{Instruction dataset example taken from OpenAI 
    grade-school-math dataset 
    }
    \label{fig:instruct_example}
\end{figure}

Instruction tuning is a technique designed to enhance the usefulness of language models by training them on Instruction/Response Tuples, as illustrated in Figure \ref{fig:instruct_example} (example taken from OpenAI grade school math dataset \cite{cobbe2021gsm8k}).  This method has demonstrated impressive capabilities across a variety of NLP tasks \cite{2020t5}. In previous work, such as T5 \cite{2020t5}, an instruction dataset is automatically crafted by leveraging various NLP tasks, including Sentiment Analysis, Paraphrasing, Natural Language Inference (NLI), Co-reference Resolution, Word Sense Disambiguation, and more. These tasks are converted into an Instruction/Answer format. For example:

\textit{“summarize: state authorities dispatched emergency crews Tuesday to survey the damage after an onslaught of severe weather in Mississippi…” –> “six people hospitalized after a storm in Attala County.”}

By assigning predetermined templates to each NLP task, such datasets can be efficiently generated.
\section{Method}

    We developed the following novel process to enable the leveraging of siloed data so we can train a generative model to generate data in the style of the original siloed data for downstream tasks: 

    \begin{figure}[t]
    \centering
    \includegraphics[width=0.48\textwidth]{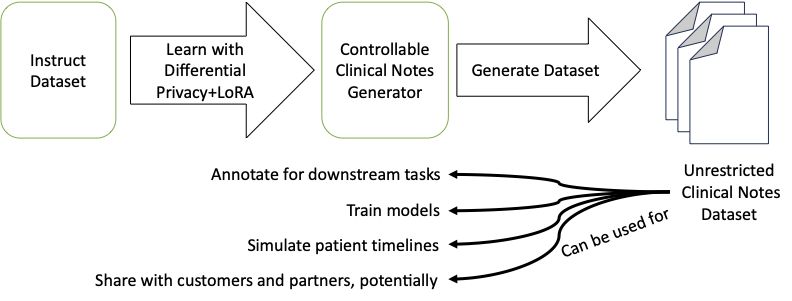}
    \caption{Tuning model over examples from the instruct dataset and various use scenarios}
    \label{fig:notes_gen}
\end{figure}
    
\begin{figure}[t]
    \centering
    \includegraphics[width=0.48\textwidth]{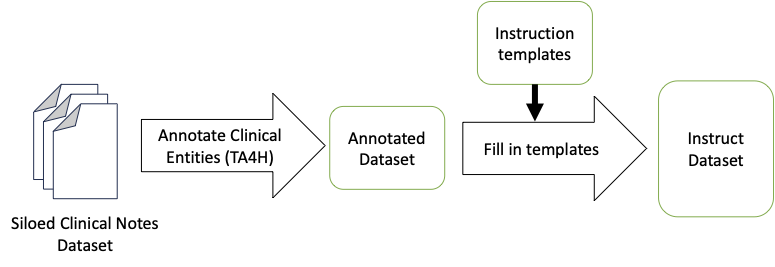}
    \caption{Instruct dataset generation
    }
    \label{fig:dataset_gen}
\end{figure}

    \begin{enumerate}
        \item Construct reusable generation instruction templates that leverage clinical structuring types (entities, relations, assertions) that can be mapped to real clinical data. For example: 
        \texttt{“\\ 
        *** ENTITIES ***\\
        - Symptom or sign: scarring\\
        - Examination name: CT chest\\
        \\ 
        *** CLINICAL NOTE OF TYPE CXR ***\\
        "\\
        }
        We used MIMIC-III documents \cite{johnson2016mimic}, 100,000 docs for training and 5,000 for training.
        \item The instructions and suitable completions are processed and populated automatically within a siloed security boundary, for example corresponding to a single medical provider's compute environment leveraging the above templates and using clinical language understanding tools (such as text analytics for health) to map entities, relations, and assertions without exposing any siloed data (see figure \ref{fig:dataset_gen}).
        \item With the siloed instruction dataset and completion pairs, we can train LLM LoRA heads \cite{hu2021lora} securely using the siloed instruction dataset, differential privacy, and anonymization methods to eliminate any privacy risks. The use of LoRA enables modeling and generation of data in the style of each silo using one base LLM model without the overhead of maintaining a new LLM model for each silo.
        \item Leveraging diverse sample prompts across a wide range of relations, entities, and assertions, a new dataset is created in the style of the original siloed data using the fine-tuned LoRA heads.
        \item Through training a model on this synthetic data, we can demonstrate that it provides similar performance metrics, when compared to a model trained on the original siloed data.
    \end{enumerate}
    This process decouples the limitations of data retention while preserving if not improving, patient privacy (see figure \ref{fig:notes_gen}).
    
\section{Results}
    \subsection{Privacy-Utility Trade-off in Trained Language Models}
    
    When training language models with Differential Privacy Stochastic Gradient Descent (DP-SGD), it is essential to evaluate the trade-off between privacy and utility. This section discusses the methods and metrics used to measure this trade-off.
    
    \subsubsection{Metrics for Privacy Evaluation}
    
    To quantify the level of privacy achieved by a trained language model, several metrics can be considered:
    
    \begin{itemize}
      \item \textbf{Epsilon ($\epsilon$) Budget}: Measures the maximum allowable privacy loss. Smaller values indicate stronger privacy guarantees.
      \item \textbf{Delta ($\delta$) Parameter}: Quantifies the probability of failing to achieve $\epsilon$-differential privacy.
      \item \textbf{Utility Metrics}: Various metrics to assess the utility of the model, such as accuracy, perplexity, or task-specific performance.
    \end{itemize}

    To evaluate the impact of privacy on model utility, we experimented with varying levels of differential privacy, selecting $\epsilon$ values of 2, 4, and 8, along with a non-private baseline. These $\epsilon$ values are commonly used in practice, striking a balance between privacy and utility; smaller values, such as 2, offer strong privacy guarantees, while larger values like 8 allow for more utility, making them suitable for comparing the trade-offs in practical applications.

    \subsubsection{Privacy Evaluation}
    Recent advances in membership inference attacks (MIAs) have demonstrated the effectiveness of the Robust Membership Inference Attack (rMIA), which significantly enhances privacy auditing capabilities while maintaining low computational costs. The rMIA method, introduced by Zarifzadeh et al. \cite{zarifzadeh2024lowcosthighpowermembershipinference}, leverages both reference models and population data to improve the likelihood ratio test used for detecting membership. In our work, due to constraints related to fine-tuning models on clinical notes, we adopted a setup similar to the rMIA's low-cost scenario, utilizing only a single target model and a single reference model (an untrained version of the target model). On a dataset of 1000 in-distribution samples, with 50\% randomly included in training, we employed black-box access to relative probabilities for calculating rMIA scores
    
    However, this setup did not provide sufficient signal to discern a clear trend between models with no privacy and those with added privacy, as all results generated an AUC around 0.5.

    These results can be attributed to several factors: the reference model used was the pre-trained model, which does not closely reflect the neighboring model expected by Differential Privacy (DP) assumptions. Furthermore, our canaries were sampled directly from the dataset distribution, as we opted against using synthetic canaries. Consequently, the constraints of our analysis precluded the deployment of a stronger adversary model that could extract more definitive signals. While the empirical privacy evaluation didn't yield clear results, we are relying on prior work proving the validity of DP-transformers\footnote{\texttt{https://github.com/microsoft/dp-transformers}}, which builds upon the well-established Opacus PyTorch library, ensuring strong differential privacy guarantees and making our approach theoretically sound.
    
    \comment{
    \subsubsection{Evaluation Methodology}
    
    The privacy and utility of trained language models are evaluated through rigorous experimentation and analysis. The following table summarizes the results obtained:
    
    \begin{table}[ht]
      \centering
      \begin{tabular}{|c|c|c|}
        \hline
        Model & $\epsilon$ & Utility Metric (e.g., Accuracy) \\
        \hline
        Model 1 & 1.0 & 85\% \\
        \hline
        Model 2 & 2.0 & 87\% \\
        \hline
        Model 3 & 4.0 & 90\% \\
        \hline
        Model 4 & 8.0 & 95\% \\
        \hline
      \end{tabular}
      \caption{Privacy and utility trade-off in trained language models}
      \label{tab:privacy-utility}
    \end{table}

    \subsubsection{Discussion}
    
    The results presented in Table~\ref{tab:privacy-utility} illustrate the trade-off between privacy (measured by $\epsilon$) and utility (measured by accuracy). \dan{which utility metric shall we use?} Models with lower $\epsilon$ values achieve higher privacy but may sacrifice some utility. Conversely, models with higher $\epsilon$ values provide better utility but at the cost of reduced privacy guarantees.
    
    In summary, evaluating the privacy-utility trade-off is crucial for designing and deploying language models that balance privacy protection with effective performance in various applications.    
    }
    \begin{table*}[t!]
    \centering
    \begin{tabular}{|l|l||l|l|l|l|}
    \hline
    MLM data Model & privacy settings & i2b2\_2009  & n2c2\_2022  & BC5CDR   & NCBI-disease \\
    \hline
    \hline
    No MLM train  & N/A             & 0.86818     & 0.81984     & 0.81391  & 0.86835      \\
    Vanilla GPT-3 & N/A             & 0.86603     & 0.82577     & 0.81722  & 0.87042      \\
    Original MIMIC data  & N/A      & \textbf{0.87517} & \textbf{0.82873} & \textbf{0.82198} & \textbf{0.88047} \\
    \hline
    GPT-3         & DP              & 0.86854     & 0.82380     & 0.81782  & \textbf{0.87751}      \\
    GPT-3         & w/o DP          & \textbf{0.87287}     & \textbf{0.83157}     & \textbf{0.82078}  & 0.87412      \\
    \hline
    PHI-2         & DP $\epsilon=2$ &         0.86931  &         0.82640  &          0.81939  &         0.86318  \\
    PHI-2         & DP $\epsilon=4$ &         0.87441  &         0.82239  &          0.81972  &         0.87092  \\
    PHI-2         & DP $\epsilon=8$ &         0.87001  & \textbf{0.82673} &          0.81749  &         0.86477  \\
    PHI-2         & w/o DP          & \textbf{0.87600} &         0.82577  &  \textbf{0.82826} & \textbf{0.87434} \\
    \hline
    \end{tabular}
    \caption{\label{table:ner}F1 values for NER tasks after training encoder models generated data, base MLM model is RoBERTa}
\end{table*}

    \subsection{Utility Test}
    \begin{figure}[t]
    \centering
    \includegraphics[width=0.48\textwidth]{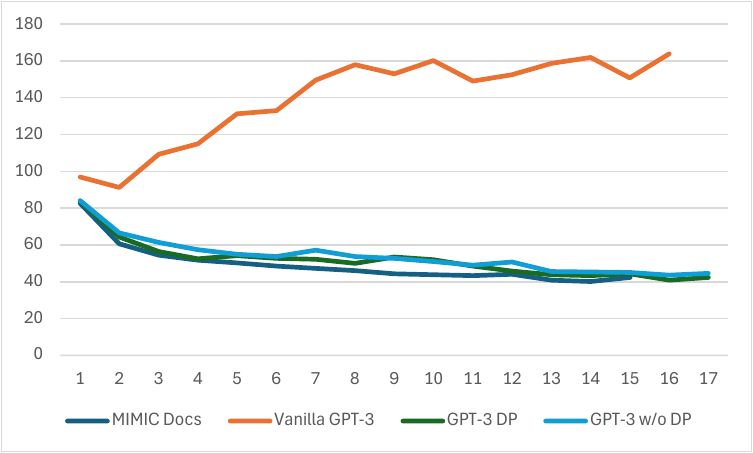}
    \includegraphics[width=0.48\textwidth]{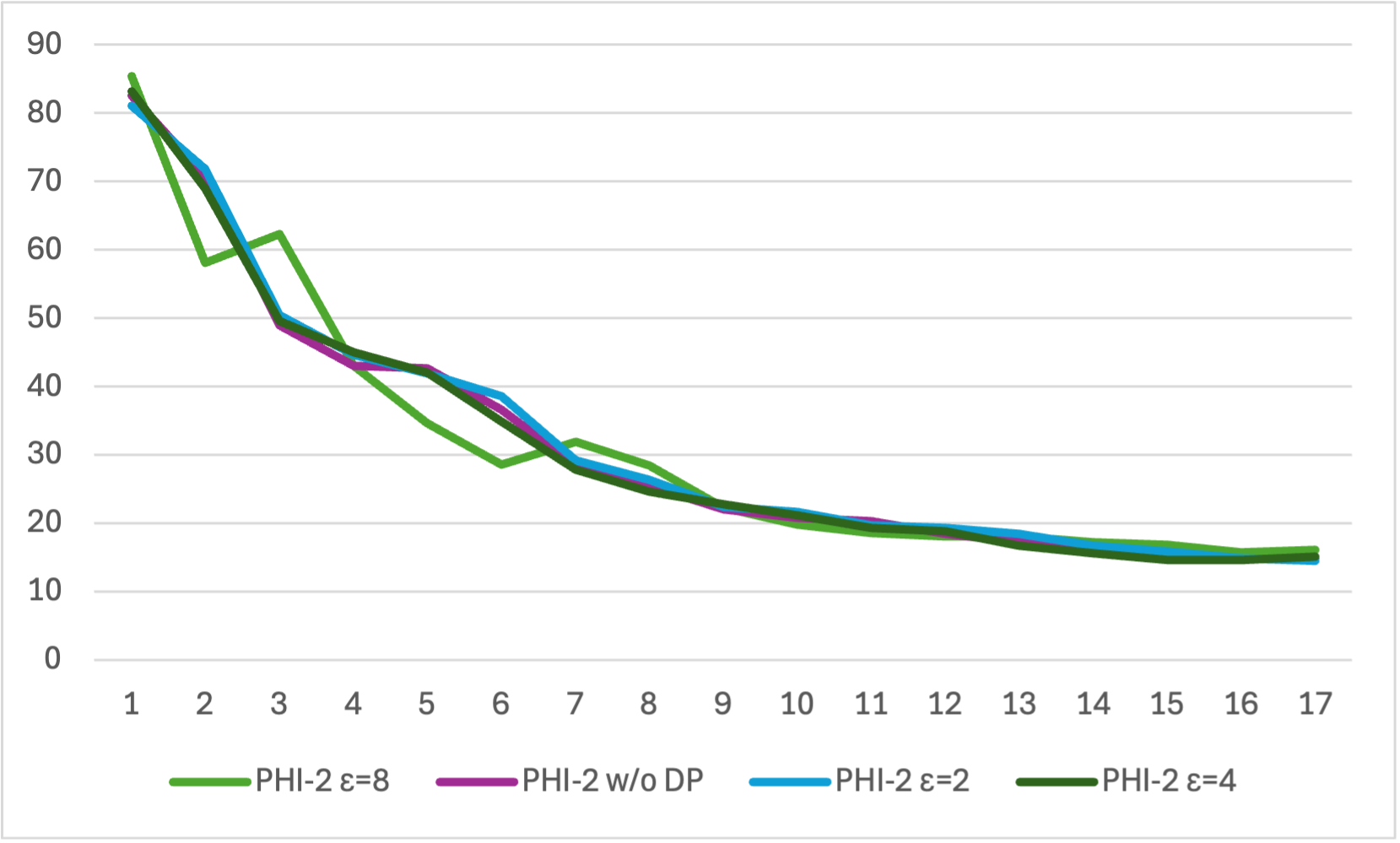}
    \caption{Perplexity over time of RoBERTa MLM train on each document set calculate against the original MIMIC texts}
    \label{fig:perplexity}
\end{figure}
    \begin{figure}[t]
    \centering
    \includegraphics[width=0.48\textwidth]{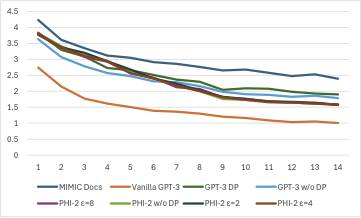}
    \caption{Train loss over time of RoBERTa MLM train on each document set}
    \label{fig:train_loss}
\end{figure}
\begin{figure}
    \centering
    \includegraphics[width=0.48\textwidth]{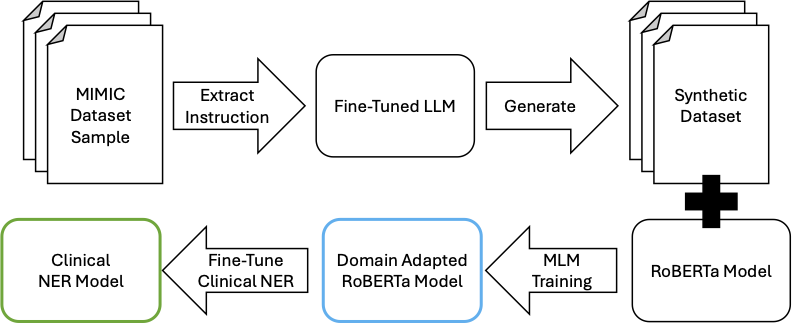}
    \caption{Process to obtain NER model for utility testing}
    \label{fig:utility}
    \vspace{-1em}
\end{figure}

    In order to evaluate the utility of various candidate models we use the following setup to measure how well their generated data can contribute to downstream tasks.  (see figure \ref{fig:utility}):
    \begin{itemize}
        \item Generate a document set using the instructions prompts created from MIMIC documents, we use a different generation LLM for each experiment
        \item Fine-tune RoBERTa \cite{liu2019roberta} masked language model (MLM) on the generated document set
        \item Train the fine-tuned RoBERTa model for NER tasks in the medical domain:
        \begin{itemize}
            \item \textbf{i2b2 2009\cite{patrick2010high}}: shared task to extract medication-related information from narrative patient records
            training (696 records) and test (533 records, 7,096 entities)
            \item \textbf{n2c2 2022\cite{mahajan2023overview}}: Social History Annotation Corpus (SHAC), this task explores the extraction of Social Determinants of Health (SDOH) from clinical notes, including substance use (alcohol, drug, tobacco), employment, and living status information [train (1,316 notes), validation (188 notes) and test set (373 notes, 3,855 entities)]
            \item \textbf{BC5CDR\cite{BCDR}}: PubMed articles with annotated chemicals, diseases and chemical-disease interactions.
            train (5228 instances), validation (5330 instances) and test set (5865 instances, 14,177 entities)
            \item \textbf{NCBI-disease\cite{dougan2014ncbi}}: PubMed abstracts annotated with disease mentions.
             train (5433 instances), validation (924 instances) and test set (941 instances, 2,055 entities)
        \end{itemize}
        
    \end{itemize}
    
    We used several language models to generate the document set: GPT-3 \cite{brown2020language} model (\texttt{code-cushman-002}) trained with DP\footnote{Due to training costs of the GPT-3 model we didn't train the model with different $\epsilon$ values}, GPT-3 model  trained without DP, Vanilla GPT-3 model, PHI-2 \cite{gunasekar2023textbooks} (chosen due to it's relatively small size) model trained with DP (with $\epsilon$ values of 2, 4 and 8) and PHI-2 model trained without DP) using a sample of instructions generated from MIMIC documents, we a compared the outputs to the original MIMIC documents as a test set for the MLM training. For future work we will evaluate generation with newer open source models such as Llama 3 \cite{dubey2024llama3herdmodels}, Mistral \cite{jiang2023mistral7b} and newer versions of Phi \cite{abdin2024phi3technicalreporthighly}.

    The hypothesis of this test is that the RoBERTa \footnote{We choose RoBeRTA as an easy to train encoder model that will enable us to show our contribution} model without any MLM training will yield the worst results since it was not adapted to the clinical domain and the best preforming model will be the one tuned on the original MIMIC document since they best represent the domain. The question we want to determine is how well our method will bridge this gap. 

    Train loss from MLM training process (Figure \ref{fig:train_loss}) illustrates how similar is the document set to the RoBERTa model expectations, while lower train loss corresponding with similarity. The most similar document set is the one generated from the vanilla GPT-3 model while the target MIMIC document set yields the highest train loss. All training process seems to converge in a similar manner over their respectful train sets.
    
    Perplexity scores from the MLM training process (Figure \ref{fig:perplexity}) show that training on documents generated by vanilla GPT-3 results in increased perplexity over the MIMIC test set, whereas all other models show a decrease. This demonstrates that our method, even with differential privacy (DP), learns to produce documents similar to the original MIMIC documents.
    
    In table \ref{table:ner} we present the weighted F1 scores of each experiment. Using the original MIMIC data to domain adapt the RoBERTa model yields the best performance while not adapting at yields the worst results. The utility of domain adapting the both GPT-3 and PHI-2 is clear while the use of DP causes only a slight drop and even improvement over w/o DP. In our setup decreasing the $\epsilon$ value seems to yield stable utility results where the best preforming $\epsilon$ value on average is 4.

\section{Conclusion}
In this work, we introduce a novel method for generating ‘clone’ datasets using domain adapted entity extraction and templated instruction tuning designed to responsibly handle ephemeral constraints. We illustrate the balance between privacy and utility achieved through our approach in the creation of such datasets. Our experiments show that while the DP training causes minor drops in performance vs the original dataset it significantly improves downstream results over not using any of the protected data and its performance is stable for different $\epsilon$ values.

\bibliography{custom}

\begin{thebibliography}{22}
\expandafter\ifx\csname natexlab\endcsname\relax\def\natexlab#1{#1}\fi

\bibitem[{Abadi et~al.(2016)Abadi, Chu, Goodfellow, McMahan, Mironov, Talwar, and Zhang}]{AbadiCGMMTZ16}
Martin Abadi, Andy Chu, Ian Goodfellow, H.~Brendan McMahan, Ilya Mironov, Kunal Talwar, and Li~Zhang. 2016.
\newblock Deep learning with differential privacy.
\newblock In \emph{Proceedings of the 2016 ACM Conference on Computer and Communications Security}, CCS '16, pages 308--318. ACM.

\bibitem[{Abdin et~al.(2024)Abdin, Jacobs, Awan, Aneja, Awadallah, Awadalla, Bach, Bahree, Bakhtiari, Bao, Behl, Benhaim, Bilenko, Bjorck, Bubeck, Cai, Cai, Mendes, Chen, Chaudhary, Chen, Chen, Chen, Chen, Chopra, Dai, Giorno, de~Rosa, Dixon, Eldan, Fragoso, Iter, Gao, Gao, Gao, Garg, Goswami, Gunasekar, Haider, Hao, Hewett, Huynh, Javaheripi, Jin, Kauffmann, Karampatziakis, Kim, Khademi, Kurilenko, Lee, Lee, Li, Li, Liang, Liden, Liu, Liu, Liu, Lin, Lin, Luo, Madan, Mazzola, Mitra, Modi, Nguyen, Norick, Patra, Perez-Becker, Portet, Pryzant, Qin, Radmilac, Rosset, Roy, Ruwase, Saarikivi, Saied, Salim, Santacroce, Shah, Shang, Sharma, Shukla, Song, Tanaka, Tupini, Wang, Wang, Wang, Wang, Ward, Wang, Witte, Wu, Wyatt, Xiao, Xu, Xu, Xu, Yadav, Yang, Yang, Yang, Yang, Yu, Yuan, Zhang, Zhang, Zhang, Zhang, Zhang, Zhang, Zhang, and Zhou}]{abdin2024phi3technicalreporthighly}
Marah Abdin, Sam~Ade Jacobs, Ammar~Ahmad Awan, Jyoti Aneja, Ahmed Awadallah, Hany Awadalla, Nguyen Bach, Amit Bahree, Arash Bakhtiari, Jianmin Bao, Harkirat Behl, Alon Benhaim, Misha Bilenko, Johan Bjorck, Sébastien Bubeck, Qin Cai, Martin Cai, Caio César~Teodoro Mendes, Weizhu Chen, Vishrav Chaudhary, Dong Chen, Dongdong Chen, Yen-Chun Chen, Yi-Ling Chen, Parul Chopra, Xiyang Dai, Allie~Del Giorno, Gustavo de~Rosa, Matthew Dixon, Ronen Eldan, Victor Fragoso, Dan Iter, Mei Gao, Min Gao, Jianfeng Gao, Amit Garg, Abhishek Goswami, Suriya Gunasekar, Emman Haider, Junheng Hao, Russell~J. Hewett, Jamie Huynh, Mojan Javaheripi, Xin Jin, Piero Kauffmann, Nikos Karampatziakis, Dongwoo Kim, Mahoud Khademi, Lev Kurilenko, James~R. Lee, Yin~Tat Lee, Yuanzhi Li, Yunsheng Li, Chen Liang, Lars Liden, Ce~Liu, Mengchen Liu, Weishung Liu, Eric Lin, Zeqi Lin, Chong Luo, Piyush Madan, Matt Mazzola, Arindam Mitra, Hardik Modi, Anh Nguyen, Brandon Norick, Barun Patra, Daniel Perez-Becker, Thomas Portet, Reid Pryzant, Heyang
  Qin, Marko Radmilac, Corby Rosset, Sambudha Roy, Olatunji Ruwase, Olli Saarikivi, Amin Saied, Adil Salim, Michael Santacroce, Shital Shah, Ning Shang, Hiteshi Sharma, Swadheen Shukla, Xia Song, Masahiro Tanaka, Andrea Tupini, Xin Wang, Lijuan Wang, Chunyu Wang, Yu~Wang, Rachel Ward, Guanhua Wang, Philipp Witte, Haiping Wu, Michael Wyatt, Bin Xiao, Can Xu, Jiahang Xu, Weijian Xu, Sonali Yadav, Fan Yang, Jianwei Yang, Ziyi Yang, Yifan Yang, Donghan Yu, Lu~Yuan, Chengruidong Zhang, Cyril Zhang, Jianwen Zhang, Li~Lyna Zhang, Yi~Zhang, Yue Zhang, Yunan Zhang, and Xiren Zhou. 2024.
\newblock \href {http://arxiv.org/abs/2404.14219} {Phi-3 technical report: A highly capable language model locally on your phone}.

\bibitem[{Bassily et~al.(2014)Bassily, Smith, and Thakurta}]{BassilyST14}
Raef Bassily, Adam Smith, and Abhradeep Thakurta. 2014.
\newblock Private empirical risk minimization: Efficient algorithms and tight error bounds.
\newblock In \emph{Proceedings of the 55th Annual IEEE Symposium on Foundations of Computer Science}, FOCS '14, pages 464--473. IEEE Computer Society.

\bibitem[{Bitran(2022)}]{ta4hblog}
Hadas Bitran. 2022.
\newblock \href {https://azure.microsoft.com/en-us/blog/expanding-ai-technology-for-unstructured-text-beyond-english/} {Expanding ai technology for unstructured biomedical text beyond english}.

\bibitem[{Bodenreider(2004)}]{bodenreider2004unified}
Olivier Bodenreider. 2004.
\newblock The unified medical language system (umls): integrating biomedical terminology.
\newblock \emph{Nucleic acids research}, 32(suppl\_1):D267--D270.

\bibitem[{Brown et~al.(2020)Brown, Mann, Ryder, Subbiah, Kaplan, Dhariwal, Neelakantan, Shyam, Sastry, Askell et~al.}]{brown2020language}
Tom Brown, Benjamin Mann, Nick Ryder, Melanie Subbiah, Jared~D Kaplan, Prafulla Dhariwal, Arvind Neelakantan, Pranav Shyam, Girish Sastry, Amanda Askell, et~al. 2020.
\newblock Language models are few-shot learners.
\newblock \emph{Advances in neural information processing systems}, 33:1877--1901.

\bibitem[{Cobbe et~al.(2021)Cobbe, Kosaraju, Bavarian, Chen, Jun, Kaiser, Plappert, Tworek, Hilton, Nakano, Hesse, and Schulman}]{cobbe2021gsm8k}
Karl Cobbe, Vineet Kosaraju, Mohammad Bavarian, Mark Chen, Heewoo Jun, Lukasz Kaiser, Matthias Plappert, Jerry Tworek, Jacob Hilton, Reiichiro Nakano, Christopher Hesse, and John Schulman. 2021.
\newblock Training verifiers to solve math word problems.
\newblock \emph{arXiv preprint arXiv:2110.14168}.

\bibitem[{Do{\u{g}}an et~al.(2014)Do{\u{g}}an, Leaman, and Lu}]{dougan2014ncbi}
Rezarta~Islamaj Do{\u{g}}an, Robert Leaman, and Zhiyong Lu. 2014.
\newblock Ncbi disease corpus: a resource for disease name recognition and concept normalization.
\newblock \emph{Journal of biomedical informatics}, 47:1--10.

\bibitem[{Dubey et~al.(2024)Dubey, Jauhri, Pandey, Kadian, Al-Dahle, Letman, Mathur, Schelten, Yang, Fan, Goyal, Hartshorn, Yang, Mitra, Sravankumar, Korenev, Hinsvark, Rao, Zhang, Rodriguez, Gregerson, Spataru, Roziere, Biron, Tang, Chern, Caucheteux, Nayak, Bi, Marra, McConnell, Keller, Touret, Wu, Wong, Ferrer, Nikolaidis, Allonsius, Song, Pintz, Livshits, Esiobu, Choudhary, Mahajan, Garcia-Olano, Perino, Hupkes, Lakomkin, AlBadawy, Lobanova, Dinan, Smith, Radenovic, Zhang, Synnaeve, Lee, Anderson, Nail, Mialon, Pang, Cucurell, Nguyen, Korevaar, Xu, Touvron, Zarov, Ibarra, Kloumann, Misra, Evtimov, Copet, Lee, Geffert, Vranes, Park, Mahadeokar, Shah, van~der Linde, Billock, Hong, Lee, Fu, Chi, Huang, Liu, Wang, Yu, Bitton, Spisak, Park, Rocca, Johnstun, Saxe, Jia, Alwala, Upasani, Plawiak, Li, Heafield, Stone, El-Arini, Iyer, Malik, Chiu, Bhalla, Rantala-Yeary, van~der Maaten, Chen, Tan, Jenkins, Martin, Madaan, Malo, Blecher, Landzaat, de~Oliveira, Muzzi, Pasupuleti, Singh, Paluri, Kardas, Oldham, Rita,
  Pavlova, Kambadur, Lewis, Si, Singh, Hassan, Goyal, Torabi, Bashlykov, Bogoychev, Chatterji, Duchenne, Çelebi, Alrassy, Zhang, Li, Vasic, Weng, Bhargava, Dubal, Krishnan, Koura, Xu, He, Dong, Srinivasan, Ganapathy, Calderer, Cabral, Stojnic, Raileanu, Girdhar, Patel, Sauvestre, Polidoro, Sumbaly, Taylor, Silva, Hou, Wang, Hosseini, Chennabasappa, Singh, Bell, Kim, Edunov, Nie, Narang, Raparthy, Shen, Wan, Bhosale, Zhang, Vandenhende, Batra, Whitman, Sootla, Collot, Gururangan, Borodinsky, Herman, Fowler, Sheasha, Georgiou, Scialom, Speckbacher, Mihaylov, Xiao, Karn, Goswami, Gupta, Ramanathan, Kerkez, Gonguet, Do, Vogeti, Petrovic, Chu, Xiong, Fu, Meers, Martinet, Wang, Tan, Xie, Jia, Wang, Goldschlag, Gaur, Babaei, Wen, Song, Zhang, Li, Mao, Coudert, Yan, Chen, Papakipos, Singh, Grattafiori, Jain, Kelsey, Shajnfeld, Gangidi, Victoria, Goldstand, Menon, Sharma, Boesenberg, Vaughan, Baevski, Feinstein, Kallet, Sangani, Yunus, Lupu, Alvarado, Caples, Gu, Ho, Poulton, Ryan, Ramchandani, Franco, Saraf,
  Chowdhury, Gabriel, Bharambe, Eisenman, Yazdan, James, Maurer, Leonhardi, Huang, Loyd, Paola, Paranjape, Liu, Wu, Ni, Hancock, Wasti, Spence, Stojkovic, Gamido, Montalvo, Parker, Burton, Mejia, Wang, Kim, Zhou, Hu, Chu, Cai, Tindal, Feichtenhofer, Civin, Beaty, Kreymer, Li, Wyatt, Adkins, Xu, Testuggine, David, Parikh, Liskovich, Foss, Wang, Le, Holland, Dowling, Jamil, Montgomery, Presani, Hahn, Wood, Brinkman, Arcaute, Dunbar, Smothers, Sun, Kreuk, Tian, Ozgenel, Caggioni, Guzmán, Kanayet, Seide, Florez, Schwarz, Badeer, Swee, Halpern, Thattai, Herman, Sizov, Guangyi, Zhang, Lakshminarayanan, Shojanazeri, Zou, Wang, Zha, Habeeb, Rudolph, Suk, Aspegren, Goldman, Damlaj, Molybog, Tufanov, Veliche, Gat, Weissman, Geboski, Kohli, Asher, Gaya, Marcus, Tang, Chan, Zhen, Reizenstein, Teboul, Zhong, Jin, Yang, Cummings, Carvill, Shepard, McPhie, Torres, Ginsburg, Wang, Wu, U, Saxena, Prasad, Khandelwal, Zand, Matosich, Veeraraghavan, Michelena, Li, Huang, Chawla, Lakhotia, Huang, Chen, Garg, A, Silva, Bell,
  Zhang, Guo, Yu, Moshkovich, Wehrstedt, Khabsa, Avalani, Bhatt, Tsimpoukelli, Mankus, Hasson, Lennie, Reso, Groshev, Naumov, Lathi, Keneally, Seltzer, Valko, Restrepo, Patel, Vyatskov, Samvelyan, Clark, Macey, Wang, Hermoso, Metanat, Rastegari, Bansal, Santhanam, Parks, White, Bawa, Singhal, Egebo, Usunier, Laptev, Dong, Zhang, Cheng, Chernoguz, Hart, Salpekar, Kalinli, Kent, Parekh, Saab, Balaji, Rittner, Bontrager, Roux, Dollar, Zvyagina, Ratanchandani, Yuvraj, Liang, Alao, Rodriguez, Ayub, Murthy, Nayani, Mitra, Li, Hogan, Battey, Wang, Maheswari, Howes, Rinott, Bondu, Datta, Chugh, Hunt, Dhillon, Sidorov, Pan, Verma, Yamamoto, Ramaswamy, Lindsay, Lindsay, Feng, Lin, Zha, Shankar, Zhang, Zhang, Wang, Agarwal, Sajuyigbe, Chintala, Max, Chen, Kehoe, Satterfield, Govindaprasad, Gupta, Cho, Virk, Subramanian, Choudhury, Goldman, Remez, Glaser, Best, Kohler, Robinson, Li, Zhang, Matthews, Chou, Shaked, Vontimitta, Ajayi, Montanez, Mohan, Kumar, Mangla, Albiero, Ionescu, Poenaru, Mihailescu, Ivanov, Li, Wang,
  Jiang, Bouaziz, Constable, Tang, Wang, Wu, Wang, Xia, Wu, Gao, Chen, Hu, Jia, Qi, Li, Zhang, Zhang, Adi, Nam, Yu, Wang, Hao, Qian, He, Rait, DeVito, Rosnbrick, Wen, Yang, and Zhao}]{dubey2024llama3herdmodels}
Abhimanyu Dubey, Abhinav Jauhri, Abhinav Pandey, Abhishek Kadian, Ahmad Al-Dahle, Aiesha Letman, Akhil Mathur, Alan Schelten, Amy Yang, Angela Fan, Anirudh Goyal, Anthony Hartshorn, Aobo Yang, Archi Mitra, Archie Sravankumar, Artem Korenev, Arthur Hinsvark, Arun Rao, Aston Zhang, Aurelien Rodriguez, Austen Gregerson, Ava Spataru, Baptiste Roziere, Bethany Biron, Binh Tang, Bobbie Chern, Charlotte Caucheteux, Chaya Nayak, Chloe Bi, Chris Marra, Chris McConnell, Christian Keller, Christophe Touret, Chunyang Wu, Corinne Wong, Cristian~Canton Ferrer, Cyrus Nikolaidis, Damien Allonsius, Daniel Song, Danielle Pintz, Danny Livshits, David Esiobu, Dhruv Choudhary, Dhruv Mahajan, Diego Garcia-Olano, Diego Perino, Dieuwke Hupkes, Egor Lakomkin, Ehab AlBadawy, Elina Lobanova, Emily Dinan, Eric~Michael Smith, Filip Radenovic, Frank Zhang, Gabriel Synnaeve, Gabrielle Lee, Georgia~Lewis Anderson, Graeme Nail, Gregoire Mialon, Guan Pang, Guillem Cucurell, Hailey Nguyen, Hannah Korevaar, Hu~Xu, Hugo Touvron, Iliyan Zarov,
  Imanol~Arrieta Ibarra, Isabel Kloumann, Ishan Misra, Ivan Evtimov, Jade Copet, Jaewon Lee, Jan Geffert, Jana Vranes, Jason Park, Jay Mahadeokar, Jeet Shah, Jelmer van~der Linde, Jennifer Billock, Jenny Hong, Jenya Lee, Jeremy Fu, Jianfeng Chi, Jianyu Huang, Jiawen Liu, Jie Wang, Jiecao Yu, Joanna Bitton, Joe Spisak, Jongsoo Park, Joseph Rocca, Joshua Johnstun, Joshua Saxe, Junteng Jia, Kalyan~Vasuden Alwala, Kartikeya Upasani, Kate Plawiak, Ke~Li, Kenneth Heafield, Kevin Stone, Khalid El-Arini, Krithika Iyer, Kshitiz Malik, Kuenley Chiu, Kunal Bhalla, Lauren Rantala-Yeary, Laurens van~der Maaten, Lawrence Chen, Liang Tan, Liz Jenkins, Louis Martin, Lovish Madaan, Lubo Malo, Lukas Blecher, Lukas Landzaat, Luke de~Oliveira, Madeline Muzzi, Mahesh Pasupuleti, Mannat Singh, Manohar Paluri, Marcin Kardas, Mathew Oldham, Mathieu Rita, Maya Pavlova, Melanie Kambadur, Mike Lewis, Min Si, Mitesh~Kumar Singh, Mona Hassan, Naman Goyal, Narjes Torabi, Nikolay Bashlykov, Nikolay Bogoychev, Niladri Chatterji, Olivier
  Duchenne, Onur Çelebi, Patrick Alrassy, Pengchuan Zhang, Pengwei Li, Petar Vasic, Peter Weng, Prajjwal Bhargava, Pratik Dubal, Praveen Krishnan, Punit~Singh Koura, Puxin Xu, Qing He, Qingxiao Dong, Ragavan Srinivasan, Raj Ganapathy, Ramon Calderer, Ricardo~Silveira Cabral, Robert Stojnic, Roberta Raileanu, Rohit Girdhar, Rohit Patel, Romain Sauvestre, Ronnie Polidoro, Roshan Sumbaly, Ross Taylor, Ruan Silva, Rui Hou, Rui Wang, Saghar Hosseini, Sahana Chennabasappa, Sanjay Singh, Sean Bell, Seohyun~Sonia Kim, Sergey Edunov, Shaoliang Nie, Sharan Narang, Sharath Raparthy, Sheng Shen, Shengye Wan, Shruti Bhosale, Shun Zhang, Simon Vandenhende, Soumya Batra, Spencer Whitman, Sten Sootla, Stephane Collot, Suchin Gururangan, Sydney Borodinsky, Tamar Herman, Tara Fowler, Tarek Sheasha, Thomas Georgiou, Thomas Scialom, Tobias Speckbacher, Todor Mihaylov, Tong Xiao, Ujjwal Karn, Vedanuj Goswami, Vibhor Gupta, Vignesh Ramanathan, Viktor Kerkez, Vincent Gonguet, Virginie Do, Vish Vogeti, Vladan Petrovic, Weiwei Chu,
  Wenhan Xiong, Wenyin Fu, Whitney Meers, Xavier Martinet, Xiaodong Wang, Xiaoqing~Ellen Tan, Xinfeng Xie, Xuchao Jia, Xuewei Wang, Yaelle Goldschlag, Yashesh Gaur, Yasmine Babaei, Yi~Wen, Yiwen Song, Yuchen Zhang, Yue Li, Yuning Mao, Zacharie~Delpierre Coudert, Zheng Yan, Zhengxing Chen, Zoe Papakipos, Aaditya Singh, Aaron Grattafiori, Abha Jain, Adam Kelsey, Adam Shajnfeld, Adithya Gangidi, Adolfo Victoria, Ahuva Goldstand, Ajay Menon, Ajay Sharma, Alex Boesenberg, Alex Vaughan, Alexei Baevski, Allie Feinstein, Amanda Kallet, Amit Sangani, Anam Yunus, Andrei Lupu, Andres Alvarado, Andrew Caples, Andrew Gu, Andrew Ho, Andrew Poulton, Andrew Ryan, Ankit Ramchandani, Annie Franco, Aparajita Saraf, Arkabandhu Chowdhury, Ashley Gabriel, Ashwin Bharambe, Assaf Eisenman, Azadeh Yazdan, Beau James, Ben Maurer, Benjamin Leonhardi, Bernie Huang, Beth Loyd, Beto~De Paola, Bhargavi Paranjape, Bing Liu, Bo~Wu, Boyu Ni, Braden Hancock, Bram Wasti, Brandon Spence, Brani Stojkovic, Brian Gamido, Britt Montalvo, Carl
  Parker, Carly Burton, Catalina Mejia, Changhan Wang, Changkyu Kim, Chao Zhou, Chester Hu, Ching-Hsiang Chu, Chris Cai, Chris Tindal, Christoph Feichtenhofer, Damon Civin, Dana Beaty, Daniel Kreymer, Daniel Li, Danny Wyatt, David Adkins, David Xu, Davide Testuggine, Delia David, Devi Parikh, Diana Liskovich, Didem Foss, Dingkang Wang, Duc Le, Dustin Holland, Edward Dowling, Eissa Jamil, Elaine Montgomery, Eleonora Presani, Emily Hahn, Emily Wood, Erik Brinkman, Esteban Arcaute, Evan Dunbar, Evan Smothers, Fei Sun, Felix Kreuk, Feng Tian, Firat Ozgenel, Francesco Caggioni, Francisco Guzmán, Frank Kanayet, Frank Seide, Gabriela~Medina Florez, Gabriella Schwarz, Gada Badeer, Georgia Swee, Gil Halpern, Govind Thattai, Grant Herman, Grigory Sizov, Guangyi, Zhang, Guna Lakshminarayanan, Hamid Shojanazeri, Han Zou, Hannah Wang, Hanwen Zha, Haroun Habeeb, Harrison Rudolph, Helen Suk, Henry Aspegren, Hunter Goldman, Ibrahim Damlaj, Igor Molybog, Igor Tufanov, Irina-Elena Veliche, Itai Gat, Jake Weissman, James
  Geboski, James Kohli, Japhet Asher, Jean-Baptiste Gaya, Jeff Marcus, Jeff Tang, Jennifer Chan, Jenny Zhen, Jeremy Reizenstein, Jeremy Teboul, Jessica Zhong, Jian Jin, Jingyi Yang, Joe Cummings, Jon Carvill, Jon Shepard, Jonathan McPhie, Jonathan Torres, Josh Ginsburg, Junjie Wang, Kai Wu, Kam~Hou U, Karan Saxena, Karthik Prasad, Kartikay Khandelwal, Katayoun Zand, Kathy Matosich, Kaushik Veeraraghavan, Kelly Michelena, Keqian Li, Kun Huang, Kunal Chawla, Kushal Lakhotia, Kyle Huang, Lailin Chen, Lakshya Garg, Lavender A, Leandro Silva, Lee Bell, Lei Zhang, Liangpeng Guo, Licheng Yu, Liron Moshkovich, Luca Wehrstedt, Madian Khabsa, Manav Avalani, Manish Bhatt, Maria Tsimpoukelli, Martynas Mankus, Matan Hasson, Matthew Lennie, Matthias Reso, Maxim Groshev, Maxim Naumov, Maya Lathi, Meghan Keneally, Michael~L. Seltzer, Michal Valko, Michelle Restrepo, Mihir Patel, Mik Vyatskov, Mikayel Samvelyan, Mike Clark, Mike Macey, Mike Wang, Miquel~Jubert Hermoso, Mo~Metanat, Mohammad Rastegari, Munish Bansal, Nandhini
  Santhanam, Natascha Parks, Natasha White, Navyata Bawa, Nayan Singhal, Nick Egebo, Nicolas Usunier, Nikolay~Pavlovich Laptev, Ning Dong, Ning Zhang, Norman Cheng, Oleg Chernoguz, Olivia Hart, Omkar Salpekar, Ozlem Kalinli, Parkin Kent, Parth Parekh, Paul Saab, Pavan Balaji, Pedro Rittner, Philip Bontrager, Pierre Roux, Piotr Dollar, Polina Zvyagina, Prashant Ratanchandani, Pritish Yuvraj, Qian Liang, Rachad Alao, Rachel Rodriguez, Rafi Ayub, Raghotham Murthy, Raghu Nayani, Rahul Mitra, Raymond Li, Rebekkah Hogan, Robin Battey, Rocky Wang, Rohan Maheswari, Russ Howes, Ruty Rinott, Sai~Jayesh Bondu, Samyak Datta, Sara Chugh, Sara Hunt, Sargun Dhillon, Sasha Sidorov, Satadru Pan, Saurabh Verma, Seiji Yamamoto, Sharadh Ramaswamy, Shaun Lindsay, Shaun Lindsay, Sheng Feng, Shenghao Lin, Shengxin~Cindy Zha, Shiva Shankar, Shuqiang Zhang, Shuqiang Zhang, Sinong Wang, Sneha Agarwal, Soji Sajuyigbe, Soumith Chintala, Stephanie Max, Stephen Chen, Steve Kehoe, Steve Satterfield, Sudarshan Govindaprasad, Sumit Gupta,
  Sungmin Cho, Sunny Virk, Suraj Subramanian, Sy~Choudhury, Sydney Goldman, Tal Remez, Tamar Glaser, Tamara Best, Thilo Kohler, Thomas Robinson, Tianhe Li, Tianjun Zhang, Tim Matthews, Timothy Chou, Tzook Shaked, Varun Vontimitta, Victoria Ajayi, Victoria Montanez, Vijai Mohan, Vinay~Satish Kumar, Vishal Mangla, Vítor Albiero, Vlad Ionescu, Vlad Poenaru, Vlad~Tiberiu Mihailescu, Vladimir Ivanov, Wei Li, Wenchen Wang, Wenwen Jiang, Wes Bouaziz, Will Constable, Xiaocheng Tang, Xiaofang Wang, Xiaojian Wu, Xiaolan Wang, Xide Xia, Xilun Wu, Xinbo Gao, Yanjun Chen, Ye~Hu, Ye~Jia, Ye~Qi, Yenda Li, Yilin Zhang, Ying Zhang, Yossi Adi, Youngjin Nam, Yu, Wang, Yuchen Hao, Yundi Qian, Yuzi He, Zach Rait, Zachary DeVito, Zef Rosnbrick, Zhaoduo Wen, Zhenyu Yang, and Zhiwei Zhao. 2024.
\newblock \href {http://arxiv.org/abs/2407.21783} {The llama 3 herd of models}.

\bibitem[{Dwork et~al.(2006{\natexlab{a}})Dwork, Kenthapadi, McSherry, Mironov, and Naor}]{DworkKMMN06}
Cynthia Dwork, Krishnaram Kenthapadi, Frank McSherry, Ilya Mironov, and Moni Naor. 2006{\natexlab{a}}.
\newblock Our data, ourselves: Privacy via distributed noise generation.
\newblock In \emph{Proceedings of the 24th Annual International Conference on the Theory and Applications of Cryptographic Techniques, EUROCRYPT '06}, pages 486--503, Berlin, Heidelberg. Springer.
\newblock 2006{\natexlab{a}}.

\bibitem[{Dwork et~al.(2006{\natexlab{b}})Dwork, McSherry, Nissim, and Smith}]{DworkMNS06}
Cynthia Dwork, Frank McSherry, Kobbi Nissim, and Adam Smith. 2006{\natexlab{b}}.
\newblock Calibrating noise to sensitivity in private data analysis.
\newblock In \emph{Proceedings of the 3rd Conference on Theory of Cryptography, TCC '06}, pages 265--284, Berlin, Heidelberg. Springer.
\newblock 2006{\natexlab{b}}.

\bibitem[{Gunasekar et~al.(2023)Gunasekar, Zhang, Aneja, Mendes, Del~Giorno, Gopi, Javaheripi, Kauffmann, de~Rosa, Saarikivi et~al.}]{gunasekar2023textbooks}
Suriya Gunasekar, Yi~Zhang, Jyoti Aneja, Caio C{\'e}sar~Teodoro Mendes, Allie Del~Giorno, Sivakanth Gopi, Mojan Javaheripi, Piero Kauffmann, Gustavo de~Rosa, Olli Saarikivi, et~al. 2023.
\newblock Textbooks are all you need.
\newblock \emph{arXiv preprint arXiv:2306.11644}.

\bibitem[{Hu et~al.(2021)Hu, Shen, Wallis, Allen-Zhu, Li, Wang, Wang, and Chen}]{hu2021lora}
Edward~J Hu, Yelong Shen, Phillip Wallis, Zeyuan Allen-Zhu, Yuanzhi Li, Shean Wang, Lu~Wang, and Weizhu Chen. 2021.
\newblock Lora: Low-rank adaptation of large language models.
\newblock \emph{arXiv preprint arXiv:2106.09685}.

\bibitem[{Jiang et~al.(2023)Jiang, Sablayrolles, Mensch, Bamford, Chaplot, de~las Casas, Bressand, Lengyel, Lample, Saulnier, Lavaud, Lachaux, Stock, Scao, Lavril, Wang, Lacroix, and Sayed}]{jiang2023mistral7b}
Albert~Q. Jiang, Alexandre Sablayrolles, Arthur Mensch, Chris Bamford, Devendra~Singh Chaplot, Diego de~las Casas, Florian Bressand, Gianna Lengyel, Guillaume Lample, Lucile Saulnier, Lélio~Renard Lavaud, Marie-Anne Lachaux, Pierre Stock, Teven~Le Scao, Thibaut Lavril, Thomas Wang, Timothée Lacroix, and William~El Sayed. 2023.
\newblock \href {http://arxiv.org/abs/2310.06825} {Mistral 7b}.

\bibitem[{Johnson et~al.(2016)Johnson, Pollard, Shen, Lehman, Feng, Ghassemi, Moody, Szolovits, Anthony~Celi, and Mark}]{johnson2016mimic}
Alistair~EW Johnson, Tom~J Pollard, Lu~Shen, Li-wei~H Lehman, Mengling Feng, Mohammad Ghassemi, Benjamin Moody, Peter Szolovits, Leo Anthony~Celi, and Roger~G Mark. 2016.
\newblock Mimic-iii, a freely accessible critical care database.
\newblock \emph{Scientific data}, 3(1):1--9.

\bibitem[{Li et~al.(2016)Li, Sun, Johnson, Sciaky, Wei, Leaman, Davis, Mattingly, Wiegers, and Lu}]{BCDR}
Jiao Li, Yueping Sun, Robin~J. Johnson, Daniela Sciaky, Chih{-}Hsuan Wei, Robert Leaman, Allan~Peter Davis, Carolyn~J. Mattingly, Thomas~C. Wiegers, and Zhiyong Lu. 2016.
\newblock \href {https://doi.org/10.1093/database/baw068} {Biocreative {V} {CDR} task corpus: a resource for chemical disease relation extraction}.
\newblock \emph{Database J. Biol. Databases Curation}, 2016.

\bibitem[{Liu et~al.(2019)Liu, Ott, Goyal, Du, Joshi, Chen, Levy, Lewis, Zettlemoyer, and Stoyanov}]{liu2019roberta}
Yinhan Liu, Myle Ott, Naman Goyal, Jingfei Du, Mandar Joshi, Danqi Chen, Omer Levy, Mike Lewis, Luke Zettlemoyer, and Veselin Stoyanov. 2019.
\newblock Roberta: A robustly optimized bert pretraining approach.
\newblock \emph{arXiv preprint arXiv:1907.11692}.

\bibitem[{Mahajan et~al.(2023)Mahajan, Liang, Tsou, and Uzuner}]{mahajan2023overview}
Diwakar Mahajan, Jennifer~J Liang, Ching-Huei Tsou, and {\"O}zlem Uzuner. 2023.
\newblock Overview of the 2022 n2c2 shared task on contextualized medication event extraction in clinical notes.
\newblock \emph{Journal of Biomedical Informatics}, 144:104432.

\bibitem[{Patrick and Li(2010)}]{patrick2010high}
Jon Patrick and Min Li. 2010.
\newblock High accuracy information extraction of medication information from clinical notes: 2009 i2b2 medication extraction challenge.
\newblock \emph{Journal of the American Medical Informatics Association}, 17(5):524--527.

\bibitem[{Raffel et~al.(2020)Raffel, Shazeer, Roberts, Lee, Narang, Matena, Zhou, Li, and Liu}]{2020t5}
Colin Raffel, Noam Shazeer, Adam Roberts, Katherine Lee, Sharan Narang, Michael Matena, Yanqi Zhou, Wei Li, and Peter~J. Liu. 2020.
\newblock \href {http://jmlr.org/papers/v21/20-074.html} {Exploring the limits of transfer learning with a unified text-to-text transformer}.
\newblock \emph{Journal of Machine Learning Research}, 21(140):1--67.

\bibitem[{Song et~al.(2013)Song, Chaudhuri, and Sarwate}]{SongCS13}
Shuang Song, Kamalika Chaudhuri, and Anand~D. Sarwate. 2013.
\newblock Stochastic gradient descent with differentially private updates.
\newblock In \emph{Proceedings of the 2013 IEEE Global Conference on Signal and Information Processing}, GlobalSIP '13, pages 245--248. IEEE Computer Society.

\bibitem[{Zarifzadeh et~al.(2024)Zarifzadeh, Liu, and Shokri}]{zarifzadeh2024lowcosthighpowermembershipinference}
Sajjad Zarifzadeh, Philippe Liu, and Reza Shokri. 2024.
\newblock \href {http://arxiv.org/abs/2312.03262} {Low-cost high-power membership inference attacks}.

\end{thebibliography}
\appendix
\section{Appendix: Compute Used Overview}
\begin{enumerate}
    \item For training GPT-3 models we used proprietary Microsoft infrastructure and cannot disclose the exact details of the compute.
    \item For PHI-2 models we use a single compute node with 8 Nvidia V100 GPU, running for 60 hours.
    \item For MLM training of RoBERTa models we used a single node with 4 Nvidia v100 running for 30 minutes.
    \item For tuned RoBERTa model training on NER task we used a single node with 4 Nvidia v100 running for 20 minutes.
\end{enumerate}

\end{document}